%% file: JRSSB.tex
\documentclass{statsoc}
\usepackage[a4paper]{geometry}
\usepackage{graphicx}
\usepackage[textwidth=8em,textsize=small]{todonotes}
\usepackage{amsmath}
\usepackage{amssymb}
\usepackage{natbib}
\newtheorem{theorem}{Theorem}
\newtheorem{lemma}{Lemma}
\newtheorem{opt}{Optimality}
\newtheorem{col}{Corollary}

\title[Spatial Process Approximations]{Spatial Process Approximations: Assessing Their Necessity}
\author[Hao Zhang]{Hao Zhang}
\address{Purdue University, West Lafayette, USA.}
\email{zhanghao@purdue.edu}

\begin{document}
\include{newcommands}
\begin{abstract}
In spatial statistics and machine learning, the kernel matrix plays a pivotal role in prediction, classification, and maximum likelihood estimation. A thorough examination reveals that for large sample sizes, the kernel matrix becomes ill-conditioned, provided the sampling locations are fairly evenly distributed. This condition poses significant challenges to numerical algorithms used in prediction and estimation computations and necessitates an approximation to prediction and the Gaussian likelihood. A review of current methodologies for managing large spatial data indicates that some fail to address this ill-conditioning problem. Such ill-conditioning often results in low-rank approximations of the stochastic processes. This paper introduces various optimality criteria and provides solutions for each.
\end{abstract}

\section{Introduction}

Spatial data collected from sampling sites across a continuous region are considered to be a partial realization of an underlying stochastic process. The ultimate objective for such data is often spatial interpolation. Linear interpolation is commonly used due to its simplicity, and Kriging is a technique for achieving the best linear unbiased interpolation which only relies on the first two moments of the stochastic process. It depends on solving linear equations that involve an $n\times n$ covariance matrix $V$, where $n$ is the number of sampling sites. The covariance matrix also appears in the Gaussian likelihood function used in both maximum likelihood estimation and Bayesian analysis of spatial data. In Kriging and statistical inferences, the inverse of the covariance matrix (also known as the precision matrix) is employed. However, advances in technology have led to sample sizes that can be thousands or even millions, making matrix inversion computationally challenging and unstable. As a result, various methods have been proposed to use approximate models to circumvent this problem.

Several methods leverage the computational efficiency of sparse matrices. One such technique is covariance tapering, which approximates $V$ using a sparse matrix \citep{furrer_covariance_2006,kaufman_covariance_2008,stein_stochastic_2013}. Studies have demonstrated that an appropriate taper leads to asymptotically optimal interpolation and efficient estimation \citep{furrer_covariance_2006,du_fixed-domain_2009}. Another approach involves using block diagonal matrices for covariance matrices \citep{stein_approximating_2004,eidsvik_estimation_2014}. Some methods alternatively employ a sparse precision matrix by assuming conditional independence  \citep{rue_approximate_2009,lindgren_explicit_2011,datta_hierarchical_2016,datta_nonseparable_2016,stroud_bayesian_2017,guinness_permutation_2018}. Other techniques utilize low-rank approximations to Gaussian processes, such as discrete process convolutions \citep{higdon_space_2002,lemos_spatio-temporal_2009}, fixed rank kriging \citep{cressie_fixed_2008,kang_bayesian_2011,katzfuss_spatio-temporal_2011}, predictive processes \citep{banerjee_gaussian_2008,finley_improving_2009}, lattice kriging \citep{nychka_multiresolution_2015}. For a review of these methods, refer to \cite{sun_geostatistics_2012}, and for comparisons, see \cite{bradley_comparison_2016} and \cite{heaton_case_2019}.

These methods were primarily designed to enhance computational efficiency and stability, especially given the challenges and computational demands of exact kriging and maximum likelihood estimation. However, our study adopts a more theoretical perspective. We provide a rigorous proof that under general conditions, both the matrix \( V \) and its inverse are ill-conditioned. This characteristic makes the computation of kriging coefficients using numerically stable algorithms nearly impossible. Merely boosting computational power is not a panacea for this problem. Therefore, approximations of either \( V \) or \( V^{-1} \) become necessary. This leads to the need for approximations of the covariance matrix or the precision matrix, as well as the search for approximate methods for Kriging and Gaussian likelihood estimation.

Our results are based on the investigation of the decay of eigenvalues of covariance matrices, which has been an understudied topic. In the case of stationary covariance, previous research has provided upper bounds and established decay rates for the eigenvalue under certain conditions. For example, \cite{schaback_error_1995} provided an upper bound for the smallest eigenvalue when the sampling sites are nearly uniformly spaced in a bounded region, while \cite{belkin_does_2019} established a decay rate in the case of Gaussian stationary covariance functions. \cite{tang_identifiability_2021} generalized the results to the Mat\'ern family. When the sampling locations are assumed to be an independent sample from some probability distribution, a central limit theorem has been established for the eigenvalues of the normalized kernel matrix \citep{koltchinskii_random_2000} and probabilistic error bounds have been established for finite samples \citep{braun_accurate_2006,jia_accurate_2009}. However, we provide an error bound for the tail sum of the eigenvalues without requiring stationarity or independence of sampling locations. Our results have potential applications beyond spatial statistics as the covariance matrix, also known as the kernel matrix in machine learning, is used in a range of fields including machine learning \citep{scholkopf_learning_2002}, computer vision \citep{bagnell_kernel_2009}, computer experiment \citep{ohagan_curve_1978,currin_bayesian_1991}, pattern recognition \citep{shawe-taylor_kernel_2004},  signal processing \citep{raykar_kernel_2007}, computational biology \citep{scholkopf_kernel_2004}, robotics \citep{deisenroth_gaussian_2015}, and pre-trained Gaussian process \citep{wang_pre-trained_2023}.

The remainder of this paper is structured as follows. In Section 2, we present the principal theorem pertaining to the eigenvalues of the kernel matrix. This main result prompts us to examine low-rank models. Consequently, in Section 3, we explore a specific construction of low-rank approximation and employ the pseudo-inverse for Kriging. In Section 4, we introduce several criteria for optimal low-rank approximation and discuss their respective solutions. The proofs for the theorems are provided in the Appendix.

\section{Eigenvalues of Kernel Matrix}

Throughout this paper, we use $H$ to denote the Hilbert space generated by $Y(s), s\in D\subset R^d$ with the inner product $(Y(s), Y(x))_H=K(s, x)$. For any $Y\in H$ and a subspace $W\subset H$, $\Pi[Y|W]$ represents the projection of $Y$ onto $W$. For $x\in R^d$, $\|x\|$ signifies the Euclidean norm. Additionally, for a subset $D\subset R^d$, $|D|$ denotes the area or volume of $D$, and $\Delta(D)=\sup\{ \|x-y\|, x, y\in D\}$ refers to the supremum of distance between any pair of points in $D$.

Before presenting the main result, we will briefly discuss the Karhunen-Loève (KL) expansion and its properties, which will provide insight into the main result and facilitate subsequent proofs. For a more comprehensive treatment of the KL expansion, we direct readers to \cite{ghanem_stochastic_1991}. Consider a second-order process $Y(s)$ with mean 0, where $s \in D$ and $D$ is a compact subset in $R^d$. Let its covariance function $K(s, x)$ be continuous in $(s, x)$. Then, $Y(s)$ can be represented as:
\begin{equation}
Y(s) = \sum_{i=1}^\infty \sqrt{\lambda_i} \varphi_i(s) \xi_i \label{KL}
\end{equation}
where $\lambda_i \ge 0$ decreases to 0 as $i \to \infty$, $\xi_i$ are uncorrelated random variables with unit variances, and the functions $\varphi_i(s)$ constitute an orthonormal system, that is,
$$
\int_D \varphi_i(s)\varphi_j(s)ds=\delta_{ij}.
$$ 

The convergence in (\ref{KL}) is in the $L_2$ sense. It follows that
	\begin{equation}
	K(s, x)=\sum_{i=1}^\infty \lambda_i \varphi_i(s)\varphi_i(x).
\label{KL-K}
	\end{equation}
which is called the Mercer's Theorem. A subsequent property is that the eigenvalues are a convergent series, i.e.,
\begin{equation} \label{eqn:sum-eigens}
    \sum_{i=1}^\infty \lambda_i=\int_D K(s, s)ds<\infty.
\end{equation}
Therefore $\sum_{i>k}\lambda_i\to 0$ if $k\to \infty$. We might approximate the process $Y(s)$ by truncating the KL expansion and the integrated mean squared error becomes
\[\int_D E\left(Y(s)-\sum_{i=1}^k \sqrt{\lambda_i}\varphi_i(s)\xi_i\right)^2 ds=\sum_{i>k}\lambda_i.\]

Indeed, the truncated KL expansion is optimal in the following sense,
	\begin{equation}
	\min \int_D E\left(Y(s)-\sum_{i=1}^k a_i(s)Z_i\right)^2 ds=\sum_{i>k}\lambda_i \label{KL-optim}
	\end{equation}
where the minimum is over all square integrable functions $a_i(s)$ and random variables $Z_i$.

The explicit convergence rates of the eigenvalues $\lambda_i$ have been studied in mathematical literature. For example, \cite{konig_eigenvalue_1986} and \cite{pietsch_eigenvalues_1987} studied convergence of eigenvalues of bounded operators on compact sets.  \cite{buhmann_approximation_2002} and \cite{santin_approximation_2016} established explicit rates for $\lambda_i$ of radial basis kernels when the spectral density satisfies some tail properties. 

Now consider that the process is observed at $n$ locations, $s_i, ~i=1, 2, \ldots, n$ and write $V_n=(K(s_i, s_j))_{i,j=1}^n$. Let $\lambda_{n,i}$ be the $i$th largest eigenvalue of $V_n$ and $\uu_i$ the corresponding eigenvector. The eigenvalues depend on $n$ obviously, and we will suppress $n$ when it does not lead to confusion. Then $\uY=(Y(s_1), Y(s_2), \ldots, Y(s_n))'$ as a random vector in $R^n$ can be written in the form of  orthogonal projection 
\begin{equation}\label{eqn:Y-decom}
\uY=\sum_{i=1}^n (\uY'\uu_i) \uu_i.
\end{equation}
It is easy to see that the terms $Z_i=\uY'\uu_i$ are uncorrelated and $\text{Var}(Z_i)=\lambda_{n,i}$. The $Z_i$'s are referred to as the principal components of $\uY$.  We can write
	\begin{equation}
	Y(s_j)=\sum_{i=1}^n\sqrt{\lambda_{n,i}}u_{ij}\xi_i\label{KL-disc}
	\end{equation}
where $\xi_i=Z_i/\sqrt{\lambda_{n,i}}$ is standardized to have a variance of 1, and $u_{ij}$ is the $i$th element of $\uu_j$. We see that (\ref{KL-disc}) is similar to (\ref{KL}). Indeed, it is referred to as the discrete KL expansion \citep[e.g.,][]{dur_optimality_1998}. There is a similar property to (\ref{KL-optim}) which we will introduce in the next section. The KL expansion and the discrete KL expansion are both special cases of the spectral theory for compact transformation in Hilbert spaces. 

Understanding the behavior of the eigenvalues \( \lambda_{n,i} \) is both practically important and intriguing. A handful of available results indicate that the ratio \( \lambda_{n,i}/n \) approaches \( \lambda_i \) as \( n \) becomes large. For instance, probabilistic error bounds for the difference \( \lambda_{n,i}/n - \lambda_i \) were provided by \cite{braun_accurate_2006} and \cite{jia_accurate_2009}. Further, \cite{belkin_approximation_2018} and \cite{tang_identifiability_2021} established upper bounds for \( \lambda_{n,i}/n \) in cases where the covariance function belongs to the Gaussian and the Matern families, respectively. These bounds are analogous to those for \( \lambda_i \).

We now state a key result on the decay of $\lambda_{n,i}$. We show  under some regularity conditions on the sampling locations $s_i$,
	\begin{equation}
	\sum_{i=k+1}^n\lambda_{n,i}/n\le c_0 \sum_{i=k+1}^n \lambda_i+o(1) \quad \mbox {  as } k, n\to \infty,\label{sum-lambda-nk}
	\end{equation}
where $c_0$ is a constant that depends on $K$ but not $n$. This leads to the following main result. 

\begin{theorem}\label{main}
Let $s_i, i=1, 2, \ldots,n$ be distinct points in a compact subset $D\in R^d$ such that the Voronoi diagram defined by 
$$D_{n,i}=\{x\in D: \|x-s_i\|=\min_{j}\|x-s_j\|\}
$$
satisfies
\begin{equation}
    \max_{1\le i\le n} \Delta(D_{n,i}) \to 0, \mbox{ and } \sup_n \frac{\max_{1\le i\le n} |D_{n,i}|}{\min_{1\le i \le n} |D_{n,i}|}<\gamma \label{D-condition}
\end{equation}
for some constant $\gamma$. 
If $K(s, x)$ is continuous in $(s, x)$,  then 
\begin{equation}
(1/n)\sum_{i=k+1}^n \lambda_{n,i}\le C_1 \sum_{i=k+1}^n \lambda_i +C_2(1/c(\delta_{\max})-1 )\label{eqn:main-01}
\end{equation}
where $C_i$ are constants that do not depend on $n$ but may depend on $K$, $\delta_{\max}=\max_i \Delta(D_{n,i})$, and the function $c(\delta)$ is defined as 
\begin{equation}
    c(\delta)=\inf\{(K(x_1, x_2)/K(x_2, x_2))^2: \|x_1-x_2\|\le \delta, x_1, x_2\in D\}.
\end{equation}
\end{theorem}

Conditions in (\ref{D-condition})  indicate that the locations become increasingly dense in $D$ and are not too unevenly scattered. The second term in the right-hand side of (\ref{eqn:main-01}) depends on the behavior of the kernel near the origin. The term necessarily reflects the effect of configuration of the locations $s_i$.  For example, if some of $s_i$ are extremely close to each other, some rows of $V_n$ will be nearly identical to each other. Consequently, $V_n$ will be nearly singular or its smallest eigenvalues are close to 0. How close they are to 0 depends on the local behavior of the kernel function. 

We can show that $\lambda_{n,1}/n$ is bounded below from 0. The next corollary immediately follows the theorem and implies that the kernel matrix $V_n$ is ill-conditioned as $n$ becomes large.

\begin{col}
Under the conditions of Theorem \ref{main}, for any $i\le n$ such that $i\to \infty$, we have
\begin{equation}
\lambda_{n,1}/\lambda_{n,i}\to \infty.\label{eqn:cond-number}
\end{equation}
\end{col}

Finally, we highlight that the theorem also indicates that the precision matrix is ill-conditioned because it shares the same condition number with the kernel matrix.  There have been a burgeoning literature on sparse approximations to the precision matrix \citep[e.g.,][]{datta_hierarchical_2016,guinness_permutation_2018,datta_sparse_2021,datta_nearest-neighbor_2022}.  It is probable that these approximations result in a matrix that is still ill-conditioned, particularly if the approximation is highly precise.  However, the focus of this work is on the approximation of the covariance matrix rather than the precision matrix.

\section{Pseudo-inverse for Spatial Prediction}

\subsection{Approximation of Kriging}
Given $n$ observations at locations $s_i$, $i=1, 2, \ldots, n$, the best linear unbiased prediction for $Y(s)$, assuming all variables have mean 0, is given by
\begin{equation}\label{kriging-pred}
\hat Y(s)=\ualpha(s)'\uY
\end{equation}
where $\ualpha(s)$ is the solution to $V_n\ualpha(s)=\uk(s)$ for
$\uk(s)=(K(s, s_i), i=1, 2, \ldots,n)'$. Due to the ill-conditioning of the matrix $V_n$, solving the linear equations can become highly sensitive to small changes in the values in $\uk(s)$, or a numerical solver for linear equations may fail to execute due to the extremely large condition number of $V_n$. In practice, this could mean that different methods such as the Gaussian elimination method, Cholesky decomposition, or QR decomposition might produce quite different results when obtaining the inverse matrix or solving linear equations. We mitigate this problem by projecting $Y(s)$ onto the
$k$-dimensional subspace generated by the first $k$ principal components, and compare it with the kriging predictor in equation (\ref{kriging-pred}).

Let $Z_i$ be the principal components of $\uY$, as in the previous section, and define
\begin{equation}
    \tilde Y(s)=\Pi[Y(s)|Z_1, Z_2, \ldots, Z_k].    
\end{equation}
Since $Z_i$'s are uncorrelated and the variance $Var(Z_i)=\lambda_{n,i}$ decreases rapidly, the main theorem implies that $Z_1, Z_2, \ldots, Z_k$ account most of the variation of $Y(s_i)$'s for a sufficiently large $k$. Therefore, we might expect that $\tilde Y(s)$ to be quite close to the Kriging predictor $\hat Y(s)$ for some $k$ much less than $n$. The following theorem quantifies the predictive performance of the two predictors. 

\begin{theorem}\label{Thm-Pseudo}
    $\tilde Y(s)$ can be expressed in terms of the $k$ eigenterms of $V_n$ as follows, 
    $$\tilde Y(s)=\uk(s)' \left(\sum_{i=1}^k \lambda_{n,i}^{-1}\uu_i\uu_i'\right)\uY.$$ Furthermore, for any $s\in D$,
    $$0<E|Y(s)-\tilde Y(s)|^2-E|Y(s)-\hat Y(s)|^2\le \|\ualpha(s)\|_2^2\; \lambda_{n,k+1}$$
    where $\ualpha(s)$ is the Kriging coefficients in (\ref{kriging-pred}) and $\|\ualpha(s)\|_2$ is the $L_2$ norm in the Euclidean space. 
\end{theorem}

There are no theoretical results about the norm of $\ualpha(s)$ in general, but our empirical studies reveal the norm is often less than or equal to 1 though this may not be always true. When the norm if bounded, the difference between the two prediction variances decays at least as fast as $\lambda_{n, k+1}$. In the case of the Mat{\'e}rn kernel, \cite{tang_identifiability_2021} has established the following upper bound:
$$\lambda_{n,k} \leq C n k^{-2 v / d-1} \mbox{ for all } k=1, \ldots, n$$
for some constant $C$. This bound can be used in the Met{\'e}rn case to determine the appropriate $k$.

We note some properties of this approximate kriging prediction. Firstly, 
$$\sum_{i=1}^n E\left(Y(s_i)-\tilde Y(s_i)\right)^2\le \sum_{i=1}^n E\left(Y(s_i)-\Pi[Y(s_i)|S_k]\right)^2$$
for any $k$-dimensional subspace $S_k$ in $H_n$, the linear space spanned by $Y(s_i), i=1, 2, \ldots,n$. We will prove this in the next section. As a result, the kriging prediction approximation demonstrates some optimal characteristic. 

Secondly, the matrix $\sum_{i=1}^k \lambda_{n,i}^{-1}\uu_i\uu_i'$ is the pseudo-inverse of the singular matrix $V_{n,k}=\sum_{i=1}^k \lambda_{n,i}\uu_i\uu_i'$. For this reason, we will refer $\tilde Y(s)$ as the pseudo-kriging. It can be viewed as substituting the kernel matrix $V_n$ with its rank-$k$ approximation, $V_{n,k}$. The Eckart-Young Theorem \citep{golub_matrix_2012} posits that $V_{n,k}$ minimizes $\|V-M\|$ for all $n\times n$ matrices $M$ of rank $k$, with the norm being either the Frobenius or $L_2$ norm. This optimal norm approximation is prevalent in machine learning.

However, to address the singularity of \(V_{n,k}\), a perturbation term is typically introduced in the existing literature. As a result, the matrix \(V_{n,k} + \tau I_n\) is used in its place, where \(\tau\) is a sufficiently small positive number. As will be detailed later, the constant \(\tau\) should be less than \(\lambda_{n,k+1}\) to ensure optimal predictive performance. Nonetheless, selecting an excessively small value for \(\tau\) might not improve the numerical stability of the matrix \(V_{n,k} + \tau I_n\). Consequently, one may just use the pseudo-kriging as an alternative. 

\subsection{Numerical Studies on Eigenvalues Decay}
Theorem \ref{Thm-Pseudo} implies that the faster $\lambda_{n,k}$ decreases in $k$, the fewer terms will be needed to achieve the desirable approximation. In general, we may expect that $\lambda_{n,i}$ to be close to $\lambda_i$ when $n$ is large and the error bounds for the difference have been established. However, for a finite $n$, $\lambda_{n,i}$ also depends on the locations $s_i$.

\begin{figure}
    \centering
    \includegraphics[width=0.5\textwidth]{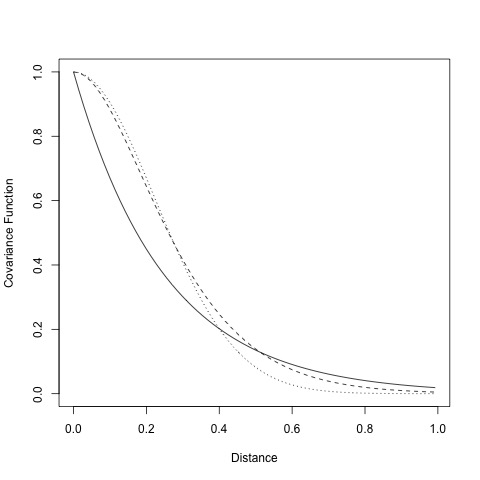}
    \caption{Plot of Three Covariance Functions ($K_1$=solid line, $K_2$=dashed line, $K_3$=dotted line).}
    \label{fig:three-matern}
\end{figure}

In this section, we run some numerical studies to examine how fast the eigenvalues $\lambda_{n,k}$ decreases as $k$ increases for a large $n$. Specifically, we examine how fast $\sum_{i=1}^k \lambda_{n,i}$ decays as $k$ increases. We first consider three covariance functions from the Mat\'ern family:
\begin{align*}
&K_1(\uh)=\exp \left(-\frac{\|\uh\|}{\phi}\right) \text{ for } \phi=0.25,\\
&K_2(\uh)=\left(1+\frac{\sqrt{5} \|\uh\|}{\phi}+\frac{5 \|\uh\|^2}{3 \phi^2}\right) \exp \left(-\frac{\sqrt{5} \|\uh\|}{\phi}\right)  \text{ for } \phi=0.25,\\
& K_3(\uh)=\exp \left( \frac{\|\uh\|^2}{\phi}\right)  \text{ for } \phi=0.1.
\end{align*}
We plot the three covariance functions in Figure \ref{fig:three-matern}. The smoothness parameter of $K_2$ is $\nu=2.5$, and the corresponding process is twice differentiable. On the other hand, $K_3$ is infinitely differentiable. It's well-known that the decay rate of the eigenvalues in the KL expansion is dependent on the smoothness parameter \citep{santin_approximation_2016}. As the smoothness parameter increases, the eigenvalues decay faster. According to Theorem \ref{main}, we anticipate that larger smoothness parameters will result in more rapid decay of the eigenvalues $\lambda_{n,k}$. 

\begin{figure}
    \centering
    \includegraphics[width=0.5\textwidth]{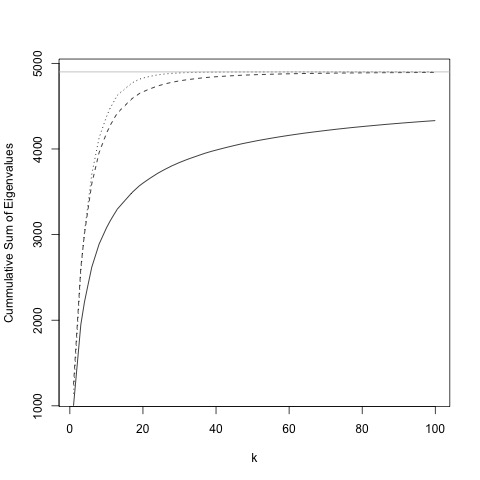}
    \caption{Cumulative Sums of Eigenvalues  ($K_1$=solid line, $K_2$=dashed line, $K_3$=dotted line)}
    \label{fig:cumsums}
\end{figure}

For each of the three kernels, we compute the covariance matrix by assuming the sampling locations are at a 70 by 70 grid in the unit square: $\{ (i/70.5, j/70.5), i,j=1, 2, \ldots, 70\}$. This results in a covariance matrix with dimensions of 4900 by 4900. We plot in Figure \ref{fig:cumsums} the cumulative sum of $\sum_{i=1}^k\lambda_{n,i}$ against $k$. If $\lambda_{n,k}$ decreases faster, the cumulative sum approaches the asymptote faster. We see that for the differentiable cases, $k\le 100$ provides sufficiently close approximation. Indeed, for $K_3$, the first 80 eigenvalues add to to 4899.995 and for $K_2$, the first 100 eigenvalues add up to 4893.675. For $K1$, the required number of eigenterms is much larger to achieve equally good approximation. When $k=500$, the sum of eigenvalues is 4657.037 or 95\% of the total variation. 

Lastly, we consider the polynomial covariance function that is used in some machine learning literature but not in spatial statistics, defined as 
$$K_4(s, x)=(1+s'x)^2, s, x\in R^2.$$ The eigenvalues decrease very rapidly, and the first 10 eigenvalues account for almost all variations.   

\begin{figure}
    \centering
    \includegraphics[width=0.5\textwidth]{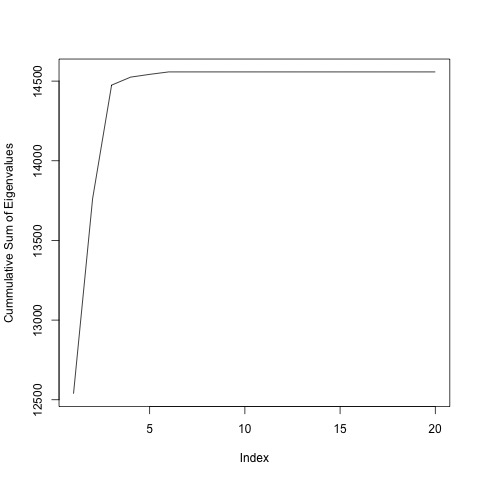}
    \caption{Cumulative Sums for the Polynomial Kernel.}
    \label{fig:polyKernel}
\end{figure}

\subsection{Numerical Approximations of Eigenvalues and Eigenvectors}

The pseudo-inverse only involves the first $k$ largest eigenvalues and eigenvectors that need to be numerically obtained when $n$ is so large that the kernel matrix is ill-conditioned. Numerical methods have been developed to approximate these eigen-terms to any desired precision, which include the random projection method through the Johnson-Lindenstrauss transformation \citep{sarlos_improved_2006,halko_finding_2011,banerjee_efficient_2013}, the Nystr\"om method \citep{drineas_nystrom_2005}, the Block Lanczos Method \citep{cullum_block_1974,golub_block_1977}, and the Block Krylov Iteration Method \citep{musco_randomized_2015}. These algorithms are able to approximate the eigenvalues and eigenvectors of an extremely large $n\times n$ matrix $V$ and can be programmed to run quite efficiently.

\subsection{The Pseudo-inverse versus the Perturbation Approach}

The optimal rank-$k$ approximation to the covariance matrix $V_n$ is obviously rank deficient if $k<n$. Theorem \ref{Thm-Pseudo} shows that the pseudo-inverse can be used for spatial prediction and compares the resulting prediction variance to the exact kriging variance. A more common practice, though, is to add a perturbation term $\tau I_n$ to the low-rank approximation \citep{williams_using_2000}. There are two situations where such a perturbation is warranted. One is when the data contain measurement error or the covariance function has a nugget effect. Another is when the Representer Theorem applies, so that the $\tau$ term represents regularization \citep{wahba_soft_2002}.

However, the nugget effect is not always present, as in some studies in computer experiments that usually use nuggetless covariance functions \citep[e.g.][]{kennedy_bayesian_2001,rasmussen_gaussian_2006}, oceanography \citep{long_aquatic_2021,jacobs_mesoscale_2001}, and climatology \citep{xu_background_2003}, among other areas. Suppose the covariance function has no nugget effect and a perturbation term is added to the rank-$k$ approximation to increase the numerical stability. Then the kriging coefficient $\ualpha$ in (\ref{kriging-pred}) is now a solution to
$$(V_{n,k}+\tau I_n)\ualpha(s)=\uk(s)$$
where $V_{n,k}=\sum_{i=1}^k \lambda_{n,i}\uu_i\uu_i'$ is the rank-$k$ approximation to $V_n$.
The interpolated value of $\uY$ is 
$$\hat \uY=V_n(V_{n,k}+\tau I_n)^{-1}\uY.$$

We can show that 
\begin{equation}\label{eqn:ESE}
    E\|\uY-\hat \uY\|^2=\sum_{i=1}^k \frac{\lambda_{n,i}}{(1+\lambda_{n,i}/\tau)^2}+\sum_{i>k}\lambda_{n,i}(1-\lambda_{n,i}/\tau)^2.
\end{equation}

The first term is strictly increasing in $\tau$ and falls within the interval $(0, \sum_{i=1}^k \lambda_{n,i})$. However, the second term decreases for $\tau < \sum_{i>k}\lambda_i^3/\sum_{i>k}\lambda_i^2<\lambda_{n,k+1}$, and then starts to increase. Consequently, the chosen $\tau$ should be less than $\lambda_{n, k+1}$ in order to minimize the sum of mean square errors. Nevertheless, the second term can become quite large for a very small $\tau$. Therefore, as we increase $\tau$ to enhance numerical stability, we might risk increasing the mean square error of prediction.

For pseudo kriging with $k$  eigenvalues, the sum of mean squared errors at the $n$ locations is $\sum_{i>k}\lambda_{n,i}$ (See Theorem \ref{Thm-Optm-C} in the next section). We now compare it that of the perturbation method  through an example. Consider the same setting as in the previous section, with the covariance function chosen to be the Gaussian $K_3$. We set $k=100$ and obtain $\sum_{i>k}\lambda_{n,i}=2.834\times 10^{-4}$. We calculate the sum of MSE in (\ref{eqn:ESE}) for four different values of $\tau$ and the corresponding condition number. Results are reported in Table \ref{Table:2}. Therefore, $\tau$ must be smaller than 0.001 in order to achieve a sum of MSE comparable to that of the pseudo-inverse method. The resulting matrix will be more ill-conditioned than when $\tau=0.001$. 

\begin{table}\label{Table:2}
\caption{Condition Numbers and Sum of MSE for the Perturbation Method}
\begin{tabular}{rrrr}
  \hline
 & $\tau$ & Condition Number & $ E\|\uY-\hat \uY\|^2$ \\ 
  \hline
  1 & 0.0010 & 1141758.43 & 0.006737 \\ 
  2 & 0.0100 & 114175.84 & 0.063977 \\ 
  3 & 0.1000 & 11417.58 & 0.602860 \\ 
  4 & 1.0000 & 1141.76 & 5.618669 \\
   \hline
\end{tabular}
\end{table}

\section{Optimal Low-Rank Approximation}

Theorem \ref{main} suggests that as $n$ increases, the covariance matrix becomes ill-conditioned, provided the sampling locations are not distributed too unevenly. This ill-conditioning issue is not limited to regular covariance matrices; it also affects tapered covariance matrices because the theorem applies to them as well. Low-rank models have been developed to address this ill-conditioning problem, and the main theorem lends further support to the use of such models. However, given the myriad of low-rank models available in the literature, it's crucial to establish criteria for selecting appropriate low-rank approximations. This section delves into that exact topic.

In this section, we explore various criteria for low-rank approximations and present the optimal solution corresponding to each criterion. Notably, these optimal solutions may not always align with what we aim to achieve in practice. Instead, they epitomize the best possible outcomes under certain conditions. Interestingly, the findings from this section will play a role in the proof for Theorem \ref{main}.

We begin by defining the low-rank approximation of a stochastic process, distinguishing it from the low-rank models used for high-dimensional data, such as principal component analysis (PCA). For a second-order process \( Y(s) \) with a continuous covariance function and mean 0, a low-rank approximation is given by:
$$
Y^*(s) = \sum_{i=1}^k a_i(s)Z_i.
$$
Here, \( Z_i \) for \( i=1,2,...,k \) represents a set of uncorrelated random variables with unit variance, while \( a_i(s) \) for \( i=1,2,...,k \) signifies real functions. Such a process is termed a rank-\( k \) approximation. For a specified \( k \), we assess the optimal rank-\( k \) process that approximates the intrinsic process \( Y(s), s \in D \). We delve into three methods to define this optimality.

\begin{opt} 
To minimize 
    $$
    \int_D E\left(Y(s)-\sum_{i=1}^k a_i(s)Z_i\right)^2 ds
    $$
over all functions real and square integrable functions $a_i(s)$ (i.e., $\int_D a_i^2(s)ds<\infty$) and all random variables $Z_i$. 
\end{opt}

We note that Optimality A is equivalent to 
$$\min_{W_k}\int_D E\left(Y(s)-\Pi[Y(s)|W_k]\right)^2 ds$$
where the minimum is over all $k$-dimensional subspace $W_k$ in $H$. The subspace $W_k$ that reaches the minimum is called the optimal subspace. The optimal subspace is spanned by the first $k$ eigenfunctions as shown in (\ref{KL-optim}).  

\begin{opt} To minimize 
    $$
    \int_D E\left(Y(s)-\sum_{i=1}^k a_i(s)Z_i\right)^2 ds
    $$
    over all real and square integrable functions $a_i(s)$  and all $Z_j\in H_n=\text{span}\{Y(s_i),~i=1, ,2\ldots,n\}$. 
\end{opt}
    
    It is equivalent to minimizing 
$$\int_D E\left(Y(s)-\Pi[Y(s)|S_k]\right)^2 ds$$
over all $k$-dimensional subspaces $S_k$ in $H_n=\text{span}\{Y(s_i),~i=1, ,2\ldots,n\}$. The optimal solution will be provided in Theorem \ref{Thm-Optm-B}.

\begin{opt}
To minimize 
    $$
    \sum_{i=1}^n E\left(Y(s_i)-\sum_{j=1}^k a_{ij}Z_j\right)^2
    $$
    over all constants $a_{ij}$ and $k$ random variables $Z_1, Z_2, \ldots, Z_k$  that are linearly independent. 
\end{opt} 

The last criterion is equivalent to minimizing 
$$\sum_{i=1}^n E\left(Y(s_i)-\Pi[Y(s_i)|S_k]\right)^2$$
over all $k$-dimensional subspaces $S_k$ in $H_n$. This criterion only concerns the process at the sampling locations $s_i$. The optimal solution will be provided in Theorem \ref{Thm-Optm-C}.

Since the optimal solution under Optimality A is known to be the truncated KL expansion, we will focus on the optimal low-rank approximations under the other two criteria.

For the underlying process $Y(s), s\in D$, define the Kriging predictor
$$Y_n^*(s)=\Pi[Y(s)|Y(s_1), Y(s_2), \ldots, Y(s_n)].$$

\cite{banerjee_gaussian_2008} called $Y_n^*(s)$ the predictive process. It turns out the the optimal solution under Optimality B can be given through the predictive process. First, we note the predictive process has a continuous covariance function when $K(s, x)$ is continuous. Hence $Y_n^*(s)$ has the Karhunen-Lo{\'e}ve expansion

\begin{equation}\label{Optim-B}
Y_n^*(s)=\sum_{i=1}^n \sqrt{\lambda_{n,i}^*} \varphi_{n,i}^*(s)Z_{ni}
\end{equation}
where the functions $\varphi_{n,i}^*(s)$ satisfy $\int_D \varphi_{n,i}^*(s)\varphi_{n,j}^*(s)ds=\delta_{ij}$, and $Z_{ni}$ are a set of uncorrelated random variables with the unit variance.  Note these $Z_{n,i}$'s generally are not in the space $H_n$.

\begin{theorem}\label{Thm-Optm-B} For $1\le k<n$, 
    \begin{equation}\label{eq:Thm-Optm-B}
 \min_{W_k} \int_D E\left(Y(s)-\Pi[Y(s)|W_k]\right)^2 ds=\sum_{i=k+1}^n \lambda_{n,i}^*+\int_D E(Y(s)-Y_n^*(s))^2ds        
    \end{equation}
where the minimum is over $k$-dimensional subspace $W_k$ in $H_n$. The optimal subspace is $W_k=\mbox{span}\{Z_{ni}^*, i=1, 2, \ldots,k\}$ and the optimal solution under Optimality B is $\sum_{i=1}^k \sqrt{\lambda_{n,i}^*} \varphi_{n,i}^*(s)Z_{ni}^*$, where $\lambda_{n,i}^*$, $\varphi_{n,i}^*$ and $Z_{n,i}^*$ are defined by (\ref{Optim-B}).
\end{theorem}

The following theorem follows results in \cite{algazi_optimality_1969} and \cite{dur_optimality_1998} and provides the solution under Optimality C. 

\begin{theorem}\label{Thm-Optm-C}
    Let $\lambda_{n,i}$ denote the $i$th eigenvalue of $V_n$, $i=1, 2, \ldots, n$. Then for any $k$-dimensional subspace $W_k\subset H$,
    $$\min_{W_k} \sum_{i=1}^n \text{E}(Y(s_i)-\Pi[Y(s_i)|W_k])^2=\sum_{i=k+1}^n \lambda_{n,i}.$$
The minimum is reached when $W_k=\text{span}\{\uu_i'\uY, i=1, 2, \ldots, k\}$ where $\uu_i$ is the eigenvector of $V_n$ corresponding to the eivenvalue $\lambda_{n,i}$. 
\end{theorem}

\section{Concluding Remarks}

We have shown that as the sample size $n$ increases, the kernel matrix tends to be ill-conditioned, provided the sampling locations are not overly dispersed. This ill-conditioning introduces numerical instability in computations related to prediction and the Gaussian likelihood. As a result, numerical approximations become imperative for the kernel matrix and its inverse.

This discovery motivates further exploration of methodologies tailored for handling extensive spatial datasets. While some techniques, like covariance tapering, fail to address the ill-conditioning problem, others, such as the low-rank approximation, prove to be effective in resolving the issue.

Such findings pave the way for intriguing questions yet to be addressed. For instance, when one employs approximations on the kernel matrix to mitigate ill-conditioning, how does this modification influence the Gaussian likelihood function? Furthermore, what ramifications does this have on the maximum likelihood estimation? We posit that these questions may be addressed through methodologies similar to those presented in \cite{tang_identifiability_2021}.

\section{Appendix}

We will first prove Theorem \ref{Thm-Optm-B} because the proof of Theorem \ref{main} depends on Theorem \ref{Thm-Optm-B}. 

\noindent \textbf{Proof of Theorem \ref{Thm-Optm-B}.} Since $W_k\subset H_n$,
$$E\left(Y(s)-\Pi[Y(s)|W_k]\right)^2=E\left(Y(s)-Y_n^*(s)\right)^2+E\left(Y^*(s)-\Pi[Y_n^*(s)|W_k]\right)^2.$$
It follows 
\begin{align*}
&\int_D E\left(Y(s)-\Pi[Y(s)|W_k]\right)^2 ds \\&=\int_D E\left(Y(s)-Y_n^*(s)\right)^2ds+ \int_D E\left(Y_n^*(s)-\Pi[Y_n^*(s)|W_k]\right)^2ds.
\end{align*}
The first term in the right hand does not depend on $W_k$, and the second term is minimized when $W_k=span\{Z_{n,i},i=1, 2, \ldots, k\}$, and the minimum is $\sum_{i>k}\lambda_{n,i}^*$ by (\ref{KL-optim}). Theorem \ref{Thm-Optm-B} is proved. 


Theorem \ref{main} is proved by taking $W_k=\text{span}\{Z_{ni}, i=1, 2, \ldots,k\}$ in both Theorems \ref{Thm-Optm-B} and \ref{Thm-Optm-C} and comparing the two minimums. Intuitively, if $\Omega$ is a small area and $s_0\in \Omega$,  
\[|\Omega| E(Y(s_0)-\Pi(Y(s_0)|W))^2\approx\int_{\Omega}E(Y(s)-\Pi[Y(s)|W] )^2ds.\]
The following lemma formally establishes the relationship. First, define
\begin{equation}
    c(\delta)=\inf\{(K(x_1, x_2)/K(x_2, x_2))^2: \|x_1-x_2\|\le \delta, x_1, x_2\in D\}.
\end{equation}
The uniform continuity of $K$ implies that $c(\delta)\to 1$ when $\delta\to 0$.

\begin{lemma}\label{Lemma1}
Let $\Omega \subset D$  and $s_0$ be a point in $\Omega$. Then for any subspace $W\subset H$,
\begin{align}
&(1/2)\int_{\Omega}E(Y(s)-\Pi[Y(s)|W] )^2ds-(1-c(\delta))\int_{\Omega}K(s,s)ds  \nonumber \\
    &\le c(\delta)|\Omega| E(Y(s_0)-\Pi(Y(s_0)|W))^2 \label{eqn:Lemma1-1}\\
    &\le 2\int_{\Omega}E(Y(s)-\Pi[Y(s)|W] )^2ds+2(1-c(\delta))\int_{\Omega}K(s,s)ds \label{eqn:Lemma1-2}
\end{align}
where $\delta=\Delta(\Omega)$.
\end{lemma}

\noindent\textbf{Proof of Lemma 1.} We regress $Y(s)$ on $Y(s_0)$ to get
$$Y(s)=\rho(s)Y(s_0)+\epsilon(s)$$
where $\rho(s)=K(s, s_0)/K(s_0, s_0)$, and
$$E(\epsilon(s)^2)=E(Y(s)^2)-\rho(s)^2E(Y(s_0)^2)\\
=K(s, s)-K(s,s_0)^2/K(s_0, s_0).
$$

It follows that 
\begin{equation}
    E\epsilon(s)^2\le K(s,s)(1-c(\delta)).\label{Eepsion}
\end{equation}
Then for any subspace $W$
\begin{align}
    & Y(s)-\Pi[Y(s)|W] \nonumber \\
    & =\rho(s)(Y(s_0)-\Pi[Y(s_0)|W])+ (\epsilon(s)-\Pi[\epsilon(s)|W]).\label{proj-ineqn1}
\end{align}
The inequality $E(X+Y)^2\le 2(EX^2+EY^2)$ for any random variables $X$ and $Y$ implies
\begin{align}
    & \rho(s)^2E(Y(s_0)-\Pi(Y(s_0)|W))^2 \nonumber\\
    &\le 2E(Y(s)-\Pi(Y(s)|W))^2+ 2E(\epsilon(s))^2, \mbox{ for any } s\in \Omega. 
\end{align}
This inequality and (\ref{Eepsion}) imply
\begin{align*}
    & \left(\int_\Omega\rho(s)^2ds \right)E(Y(s_0)-\Pi(Y(s_0)|W))^2 \nonumber\\
    &\le 2\int_\Omega E(Y(s)-\Pi(Y(s)|W))^2+ 2(1-c(\delta))\int_D K(s, s)ds. 
\end{align*}
The inequality (\ref{eqn:Lemma1-2}) follows because $\rho(s)^2\ge c(\delta)$. Equation (\ref{eqn:Lemma1-1}) can be proved similarly by applying the inequality $E(X^2)\ge (1/2)E(X-Y)^2-E(Y^2)$ for any two random variables $X$ and $Y$ to Eq. (\ref{proj-ineqn1}).

\begin{lemma} Let $\lambda_{n,i}^*$ be the eigenvalues in (\ref{Optim-B}) and $\lambda_i$ the eigenvalue in (\ref{KL}).
    For any $i\le n$, $\lambda_{n,i}^*\le \lambda_i$.
\end{lemma}

\noindent \textbf{Proof.} The Lemma follows Proposition 5 in \cite{santin_approximation_2016}.

\noindent \textbf{Proof of Theorem \ref{main}.} For simplicity, we suppress $n$ in $D_{n,i}$ and other notations when no confusions arise. Define $\delta_i=\Delta(D_i)$. By Lemma \ref{Lemma1}, 
\begin{align}
    &c(\delta_i)|D_i| E(Y(s_i)-\Pi(Y(s_i)|W_k))^2 \nonumber \\
    &\le 2\int_{D_i}E(Y(s)-\Pi[Y(s)|W_k] )^2ds+(1-c(\delta_i))\int_{D_i}K(s,s)ds. \label{eqn:integrals}
\end{align}
Since 
$
|D|=\sum_{i=1}^n |D_i|\le n \max |D_i|\le n\gamma \min_i |D_i|,$ we get
$$\min_i |D_i|\ge |D|/(n\gamma).$$

Using the last inequality and adding up both sides of (\ref{eqn:integrals}), we see
\begin{align*}
    & \min_i c(\delta_i)(n\gamma)^{-1}|D|\sum_{i=1}^nE(Y(s_i)-\Pi(Y(s_i)|W_k))^2\\
    &\le 2\int_{D}E(Y(s)-\Pi[Y(s)|W_k] )^2ds+(1-\min_i c(\delta_i))\int_{D}K(s,s)ds
\end{align*}
The inequality is true for any $k$-dimensional space $W_k$. In particular, taking $W_k$ to the be optimal subspace in Theorem \ref{Thm-Optm-B} and applying Theorem \ref{Thm-Optm-C}, we have
\begin{align*}
    \min_i c(\delta_i)(n\gamma)^{-1}|D|\sum_{i>k}\lambda_{n,i}&\le 2\left(\sum_{i=k+1}^n \lambda_{n,i}^*+\int_D E(Y(s)-Y_n^*(s))^2ds\right)\\
    &+(1-\min_i c(\delta_i) )\int_{D}K(s,s)ds\\
    &\le 2\left(\sum_{i>k}\lambda_i+\int_D E(Y(s)-Y_n^*(s))^2ds\right)\\
        &+(1-\min_i c(\delta_i) )\int_{D}K(s,s)ds.
\end{align*}
The last inequality follows Lemma 2. Let us bound the integral now.
\begin{align}
& \int_D E(Y(s)-Y_n^*(s))^2ds= \sum_{i=1}^n \int_{D_i} E(Y(s)-Y_n^*(s))^2ds\\
&\le \sum_{i=1}^n \int_{D_i} E(Y(s)-\Pi[Y(s)|Y(s_i)])^2ds .
\end{align}
Applying (\ref{proj-ineqn1}), for $s\in D_i$, we get
$$
E(Y(s)-\Pi[Y(s)|Y(s_i)])^2\le K(s_i, s_i) (1-c(\delta_i)).
$$
It follows 
$$
\int_D E(Y(s)-Y_n^*(s))^2ds \le |D|(1-\min_i c(\delta_i))\max_{s\in D} K(s, s).
$$
We have established 
$$
\min_i c(\delta_i)(n\gamma)^{-1}|D|\sum_{i>k}\lambda_{n,i}\le 2\sum_{i>k} \lambda_i+(1+|D|)(1-\min_i c(\delta_i))\max_{s\in D} K(s, s).
$$
The theorem follows.



\noindent\textbf{Proof of Corollary 1. } From (\ref{eqn:main-01}), it is seen that $\lim_{k\to \infty}\sum_{i>k}\lambda_{n,i}/n\to 0$. Therefore for any $i\le n$ and $i\to\infty$, $\lambda_{n,i}/n\to 0$. The corollary holds if $\liminf_{n\to \infty}\lambda_{n, 1}/n>0$. Therefore, it suffices to show
\begin{equation}\label{eqn:lambda_n-limit}
\lambda_{n, 1}/n\ge a \lambda_1+o(1) \mbox{ as } n\to \infty.
\end{equation}

To that end, we first observe that Mercer's Theorem (\ref{KL-K}) implies
\begin{equation*}
\lambda_1 =\int_D \int_D \phi_1(x) K(x, y) \phi_1(y) d x d y.
\end{equation*}

The continuity of the integrant implies that the integral can be approximated by a finite sum, i.e., 
\begin{equation*}
\lambda_1 =\sum_{i=1}^n\sum_{j=1}^n |D_i| |D_j|\phi_1(s_i)\phi_1(s_j)K(s_i, s_j)+o(1)
\end{equation*}

Similarly,
\begin{equation*}
1=\int_D \phi_1^2(x) dx=\sum_{i=1}^n|D_i|\phi_1^2(s_i)+o(1)
\end{equation*}

It follows the last equations that
\begin{equation}\label{eqn:lambda1}
\lambda_1=\frac{\sum_{i=1}^n\sum_{j=1}^n |D_i| |D_j|\phi_1(s_i)\phi_1(s_j)K(s_i, s_j)}{\sum_{i=1}^n|D_i|\phi_1^2(s_i)}+o(1).
\end{equation}

As we previously established, $\min_i |D_i|\ge  |D|/(rn)$. Then


Then, from (\ref{eqn:lambda1}), 
\begin{align*}
\lambda_1 &\le \frac{|D|}{rn}\frac{\sum_{i=1}^n\sum_{j=1}^n |D_i||D_j|\phi_1(s_i)\phi_1(s_j)K(s_i, s_j)}{\sum_{i=1}^n D_i^2\phi_1^2(s_i)}+o(1) \\
& \le \frac{|D|}{rn} \max_{(x_1, x_2, \ldots, x_n)\in R^n} \frac{\sum_{i=1}^n\sum_{j=1}^n x_ix_jK(s_i, s_j)}{\sum_{i=1}^n x_i^2}+o(1)\\
&= \frac{|D|}{rn}\lambda_{n,1}+o(1).
\end{align*}
Then (\ref{eqn:lambda_n-limit}) immediately follows. The corollary is proved. 

\noindent \textbf{Proof of Equation (\ref{eqn:ESE}).}
First, note the matrix $V_{n,k}+\tau I_n$ has eigenvalues $w_i=\lambda_{n,i}+\tau$ for $i\le k$ and $w_i=\tau$ for $i>k$, and shares the same eigenvectors $\uu_i$ with $V_n$. Then 
\begin{align*}
    & (V_{n,k}+\tau I_n)^{-1}=\sum_{i=1}^n w_i^{-1}\uu_i\uu_i'\\
    & V_n(V_{n,k}+\tau I_n)^{-1}=\sum_{i=1}^n \lambda_{n,i}w_i^{-1}\uu_i\uu_i'.
\end{align*}

Hence $\hat \uY=\sum_{i=1}^n \lambda_{n,i}w_i^{-1}Z_i\uu_i$ for $Z_i=\uu_i'\uY$. Applying (\ref{eqn:Y-decom}) to get
$$\uY-\hat \uY=\sum_{i=1}^n (1-\lambda_{n,i}w_i^{-1})Z_i\uu_i$$
and 
$$\|\uY-\hat \uY\|^2=\sum_{i=1}^n (1-\lambda_{n,i}w_i^{-1})^2 Z_i^2.$$ 
Equation (\ref{eqn:ESE}) is proved by observing $EZ_i^2=\lambda_{n,i}$ and $Z_i$'s are uncorrelated.








\end{document}

%% file: newcommands.tex
\newcommand{\bs}{\boldsymbol}
\newcommand{\ba}    {\begin{array}}
\newcommand{\ea}    {\end{array}}
\newcommand{\be}    {\begin{equation}}
\newcommand{\ee}    {\end{equation}}
\newcommand{\bea}    {\begin{eqnarray}}
\newcommand{\eea}    {\end{eqnarray}}
\newcommand{\nn}     {\nonumber}
\def\refhg{\hangindent=20pt\hangafter=1}
\def\refmark{\par\vskip 2mm\noindent\refhg}
\newcommand{\Var}{\operatorname{Var}}
\newcommand{\Cov}{\operatorname{Cov}}
\newcommand{\E}  {\mbox{E}}
\newcommand{\uA}       {\mbox{\boldmath$A$}}
\newcommand{\ua}       {\mbox{\boldmath$a$}}
\newcommand{\uB}       {\mbox{\boldmath$B$}}
\newcommand{\ub}       {\mbox{\boldmath$b$}}
\newcommand{\uC}       {\mbox{\boldmath$C$}}
\newcommand{\uc}       {\mbox{\boldmath$c$}}
\newcommand{\uD}       {\mbox{\boldmath$D$}}
\newcommand{\ud}       {\mbox{\boldmath$d$}}
\newcommand{\uE}       {\mbox{\boldmath$E$}}
\newcommand{\ue}       {\mbox{\boldmath$e$}}
\newcommand{\uF}       {\mbox{\boldmath$F$}}
\newcommand{\uf}       {\mbox{\boldmath$f$}}
\newcommand{\uG}       {\mbox{\boldmath$G$}}
\newcommand{\ug}       {\mbox{\boldmath$g$}}
\newcommand{\uH}       {\mbox{\boldmath$H$}}
\newcommand{\uh}       {\mbox{\boldmath$h$}}
\newcommand{\uI}       {\mbox{\boldmath$I$}}
\newcommand{\ui}       {\mbox{\boldmath$i$}}
\newcommand{\uJ}       {\mbox{\boldmath$J$}}
\newcommand{\uj}       {\mbox{\boldmath$j$}}
\newcommand{\uK}       {\mbox{\boldmath$K$}}
\newcommand{\uk}       {\mbox{\boldmath$k$}}
\newcommand{\uL}       {\mbox{\boldmath$L$}}
\newcommand{\ul}       {\mbox{\boldmath$l$}}
\newcommand{\uM}       {\mbox{\boldmath$M$}}
\newcommand{\um}       {\mbox{\boldmath$m$}}
\newcommand{\uN}       {\mbox{\boldmath$N$}}
\newcommand{\un}       {\mbox{\boldmath$n$}}
\newcommand{\uO}       {\mbox{\boldmath$O$}}
\newcommand{\uP}       {\mbox{\boldmath$P$}}
\newcommand{\up}       {\mbox{\boldmath$p$}}
\newcommand{\uQ}       {\mbox{\boldmath$Q$}}
\newcommand{\uq}       {\mbox{\boldmath$q$}}
\newcommand{\uR}       {\mbox{\boldmath$R$}}
\newcommand{\ur}       {\mbox{\boldmath$r$}}
\newcommand{\uS}       {\mbox{\boldmath$S$}}
\newcommand{\us}       {\mbox{\boldmath$s$}}
\newcommand{\uT}       {\mbox{\boldmath$T$}}
\newcommand{\ut}       {\mbox{\boldmath$t$}}
\newcommand{\uU}       {\mbox{\boldmath$U$}}
\newcommand{\uu}       {\mbox{\boldmath$u$}}
\newcommand{\uV}       {\mbox{\boldmath$V$}}
\newcommand{\uv}       {\mbox{\boldmath$v$}}
\newcommand{\uW}       {\mbox{\boldmath$W$}}
\newcommand{\uw}       {\mbox{\boldmath$w$}}
\newcommand{\uX}       {\mbox{\boldmath$X$}}
\newcommand{\ux}       {\mbox{\boldmath$x$}}
\newcommand{\uY}       {\mbox{\boldmath$Y$}}
\newcommand{\uy}       {\mbox{\boldmath$y$}}
\newcommand{\uZ}       {\mbox{\boldmath$Z$}}
\newcommand{\uz}       {\mbox{\boldmath$z$}}
\newcommand{\ualpha}            {\mbox{\boldmath$\alpha$}}
\newcommand{\ubeta}             {\mbox{\boldmath$\beta$}}
\newcommand{\ugamma}            {\mbox{\boldmath$\gamma$}}
\newcommand{\udelta}            {\mbox{\boldmath$\delta$}}
\newcommand{\uepsilon}          {\mbox{\boldmath$\epsilon$}}
\newcommand{\uvarepsilon}       {\mbox{\boldmath$\varepsilon$}}
\newcommand{\uzeta}             {\mbox{\boldmath$\zeta$}}
\newcommand{\ueta}              {\mbox{\boldmath$\eta$}}
\newcommand{\utheta}            {\mbox{\boldmath$\theta$}}
\newcommand{\uvartheta}         {\mbox{\boldmath$\vartheta$}}
\newcommand{\uiota}             {\mbox{\boldmath$\uiota$}}
\newcommand{\ukappa}            {\mbox{\boldmath$\kappa$}}
\newcommand{\ulambda}           {\mbox{\boldmath$\lambda$}}
\newcommand{\umu}               {\mbox{\boldmath$\mu$}}
\newcommand{\unu}               {\mbox{\boldmath$\nu$}}
\newcommand{\uxi}               {\mbox{\boldmath$\xi$}}
\newcommand{\uo}                {\mbox{\boldmath$\o$}}
\newcommand{\upi}               {\mbox{\boldmath$\pi$}}
\newcommand{\uvarpi}            {\mbox{\boldmath$\varpi$}}
\newcommand{\urho}              {\mbox{\boldmath$\rho$}}
\newcommand{\uvarrho}           {\mbox{\boldmath$\varrho$}}
\newcommand{\usigma}            {\mbox{\boldmath$\sigma$}}
\newcommand{\uvarsigma}         {\mbox{\boldmath$\varsigma$}}
\newcommand{\utau}              {\mbox{\boldmath$\tau$}}
\newcommand{\uupsilon}          {\mbox{\boldmath$\upsilon$}}
\newcommand{\uphi}              {\mbox{\boldmath$\phi$}}
\newcommand{\uvarphi}           {\mbox{\boldmath$\varphi$}}
\newcommand{\uchi}              {\mbox{\boldmath$\chi$}}
\newcommand{\upsi}              {\mbox{\boldmath$\psi$}}
\newcommand{\uomega}            {\mbox{\boldmath$\omega$}}
\newcommand{\uGamma}            {\mbox{\boldmath$\Gamma$}}
\newcommand{\uDelta}            {\mbox{\boldmath$\Delta$}}
\newcommand{\uTheta}            {\mbox{\boldmath$\Theta$}}
\newcommand{\uLambda}           {\mbox{\boldmath$\Lambda$}}
\newcommand{\uXi}               {\mbox{\boldmath$\Xi$}}
\newcommand{\uPi}                {\mbox{\boldmath$\Pi$}}
\newcommand{\uSigma}            {\mbox{\boldmath$\Sigma$}}
\newcommand{\uUpsilon}          {\mbox{\boldmath$\Upsilon$}}
\newcommand{\uPhi}              {\mbox{\boldmath$\Phi$}}
\newcommand{\uPsi}              {\mbox{\boldmath$\Psi$}}
\newcommand{\uOmega}            {\mbox{\boldmath$\Omega$}}
\newcommand{\uzero}            {\mbox{\boldmath$0$}}
\newcommand{\uone}               {\mbox{\boldmath$1$}}

%% file: JRSSB.bbl
\begin{thebibliography}{64}
\expandafter\ifx\csname natexlab\endcsname\relax\def\natexlab#1{#1}\fi
\expandafter\ifx\csname url\endcsname\relax
  \def\url#1{\texttt{#1}}\fi
\expandafter\ifx\csname urlprefix\endcsname\relax\def\urlprefix{URL: }\fi

\bibitem[{Algazi and Sakrison(1969)}]{algazi_optimality_1969}
Algazi, V. and Sakrison, D. (1969) On the optimality of the {Karhunen}-{Loève}
  expansion ({Corresp}.).
\newblock \textit{IEEE Transactions on Information Theory}, \textbf{15},
  319--321.

\bibitem[{Bagnell et~al.(2009)Bagnell, Bradley and
  Hebert}]{bagnell_kernel_2009}
Bagnell, J.~A., Bradley, D. and Hebert, M. (2009) Kernel methods in computer
  vision.
\newblock \textit{Foundations and Trends in Computer Graphics and Vision},
  \textbf{4}, 1--196.
\newblock Publisher: Now Publishers Inc.

\bibitem[{Banerjee et~al.(2013)Banerjee, Dunson and
  Tokdar}]{banerjee_efficient_2013}
Banerjee, A., Dunson, D.~B. and Tokdar, S.~T. (2013) Efficient {Gaussian}
  process regression for large datasets.
\newblock \textit{Biometrika}, \textbf{100}, 75--89.
\newblock \urlprefix\url{http://www.jstor.org/stable/43304538}.
\newblock Publisher: [Oxford University Press, Biometrika Trust].

\bibitem[{Banerjee et~al.(2008)Banerjee, Gelfand, Finley and
  Sang}]{banerjee_gaussian_2008}
Banerjee, S., Gelfand, A.~E., Finley, A.~O. and Sang, H. (2008) Gaussian
  predictive process models for large spatial data sets.
\newblock \textit{Journal of the Royal Statistical Society: Series B
  (Statistical Methodology)}, \textbf{70}, 825--848.
\newblock
  \urlprefix\url{https://onlinelibrary.wiley.com/doi/abs/10.1111/j.1467-9868.2008.00663.x}.
\newblock \_eprint:
  https://onlinelibrary.wiley.com/doi/pdf/10.1111/j.1467-9868.2008.00663.x.

\bibitem[{Belkin(2018)}]{belkin_approximation_2018}
Belkin, M. (2018) Approximation beats concentration? {An} approximation view on
  inference with smooth radial kernels.
\newblock \textit{Proceedings of Machine Learning Research}, \textbf{75},
  1--14.

\bibitem[{Belkin et~al.(2019)Belkin, Rakhlin and Tsybakov}]{belkin_does_2019}
Belkin, M., Rakhlin, A. and Tsybakov, A.~B. (2019) Does data interpolation
  contradict statistical optimality?
\newblock In \textit{Proceedings of the {Twenty}-{Second} {International}
  {Conference} on {Artificial} {Intelligence} and {Statistics}} (eds.
  K.~Chaudhuri and M.~Sugiyama), vol.~89 of \textit{Proceedings of {Machine}
  {Learning} {Research}}, 1611--1619. PMLR.
\newblock \urlprefix\url{https://proceedings.mlr.press/v89/belkin19a.html}.

\bibitem[{Bradley et~al.(2016)Bradley, Cressie and
  Shi}]{bradley_comparison_2016}
Bradley, J.~R., Cressie, N. and Shi, T. (2016) A comparison of spatial
  predictors when datasets could be very large.
\newblock \textit{Statistics Surveys}, \textbf{10}, 100--131.
\newblock Publisher: Amer. Statist. Assoc., the Bernoulli Soc., the Inst. Math.
  Statist., and the Statist. Soc. Canada.

\bibitem[{Braun(2006)}]{braun_accurate_2006}
Braun, M.~L. (2006) Accurate {Error} {Bounds} for the {Eigenvalues} of the
  {Kernel} {Matrix}.
\newblock \textit{Journal of Machine Learning Research}, \textbf{7},
  2303--2328.
\newblock \urlprefix\url{http://jmlr.org/papers/v7/braun06a.html}.

\bibitem[{Cressie and Johannesson(2008)}]{cressie_fixed_2008}
Cressie, N. and Johannesson, G. (2008) Fixed rank kriging for very large
  spatial data sets.
\newblock \textit{Journal of the Royal Statistical Society: Series B
  (Statistical Methodology)}, \textbf{70}, 209--226.
\newblock Publisher: Wiley.

\bibitem[{Cullum and Donath(1974)}]{cullum_block_1974}
Cullum, J. and Donath, W.~E. (1974) A block {Lanczos} algorithm for computing
  the q algebraically largest eigenvalues and a corresponding eigenspace of
  large, sparse, real symmetric matrices.
\newblock In \textit{{IEEE} {Conference} on {Decision} and {Control} including
  the 13th {Symposium} on {Adaptive} {Processes}}, 505--509.

\bibitem[{Currin et~al.(1991)Currin, Mitchell, Morris and
  Ylvisaker}]{currin_bayesian_1991}
Currin, C., Mitchell, T., Morris, M. and Ylvisaker, D. (1991) Bayesian
  {Prediction} of {Deterministic} {Functions}, with {Applications} to the
  {Design} and {Analysis} of {Computer} {Experiments}.
\newblock \textit{Journal of the American Statistical Association},
  \textbf{86}, 953--963.
\newblock \urlprefix\url{https://www.jstor.org/stable/2290511}.

\bibitem[{Datta(2021)}]{datta_sparse_2021}
Datta, A. (2021) Sparse nearest neighbor {Cholesky} matrices in spatial
  statistics.
\newblock \urlprefix\url{http://arxiv.org/abs/2102.13299}.
\newblock ArXiv:2102.13299 [stat].

\bibitem[{Datta(2022)}]{datta_nearest-neighbor_2022}
--- (2022) Nearest-neighbor sparse {Cholesky} matrices in spatial statistics.
\newblock \textit{WIREs Computational Statistics}, \textbf{14}, e1574.
\newblock
  \urlprefix\url{https://onlinelibrary.wiley.com/doi/abs/10.1002/wics.1574}.
\newblock \_eprint: https://onlinelibrary.wiley.com/doi/pdf/10.1002/wics.1574.

\bibitem[{Datta et~al.(2016{\natexlab{a}})Datta, Banerjee, Finley and
  Gelfand}]{datta_hierarchical_2016}
Datta, A., Banerjee, S., Finley, A.~O. and Gelfand, A.~E. (2016{\natexlab{a}})
  Hierarchical {Nearest}-{Neighbor} {Gaussian} {Process} {Models} for {Large}
  {Geostatistical} {Datasets}.
\newblock \textit{Journal of the American Statistical Association},
  \textbf{111}, 800--812.
\newblock \urlprefix\url{https://doi.org/10.1080/01621459.2015.1044091}.
\newblock Publisher: Taylor \& Francis \_eprint:
  https://doi.org/10.1080/01621459.2015.1044091.

\bibitem[{Datta et~al.(2016{\natexlab{b}})Datta, Banerjee, Finley, Hamm, Schaap
  and al}]{datta_nonseparable_2016}
Datta, A., Banerjee, S., Finley, A.~O., Hamm, N.~A., Schaap, M. and al, e.
  (2016{\natexlab{b}}) Nonseparable dynamic nearest neighbor {Gaussian} process
  models for large spatio-temporal data with an application to particulate
  matter analysis.
\newblock \textit{The Annals of Applied Statistics}, \textbf{10}, 1286--1316.
\newblock Publisher: Institute of Mathematical Statistics.

\bibitem[{Deisenroth et~al.(2015)Deisenroth, Fox and
  Rasmussen}]{deisenroth_gaussian_2015}
Deisenroth, M.~P., Fox, D. and Rasmussen, C.~E. (2015) Gaussian {Processes} for
  {Data}-{Efficient} {Learning} in {Robotics} and {Control}.
\newblock \textit{IEEE Transactions on Pattern Analysis and Machine
  Intelligence}, \textbf{37}, 408--423.
\newblock \urlprefix\url{http://arxiv.org/abs/1502.02860}.
\newblock ArXiv:1502.02860 [cs, stat].

\bibitem[{Drineas and Mahoney(2005)}]{drineas_nystrom_2005}
Drineas, P. and Mahoney, M.~W. (2005) On the {Nystrom} {Method} for
  {Approximating} a {Gram} {Matrix} for {Improved} {Kernel}-{Based} {Learning}.
\newblock \textit{Journal of Machine Learning Research}, \textbf{6},
  2153--2175.

\bibitem[{D\"ur(1998)}]{dur_optimality_1998}
D\"ur, A. (1998) On the {Optimality} of the {Discrete}
  {Karhunen}–{Loève} {Expansion}.
\newblock \textit{SIAM Journal on Control and Optimization}, \textbf{36},
  1937--1939.

\bibitem[{Du et~al.(2009)Du, Zhang and Mandrekar}]{du_fixed-domain_2009}
Du, J., Zhang, H. and Mandrekar, V.~S. (2009) Fixed-domain asymptotic
  properties of tapered maximum likelihood estimators.
\newblock \textit{Annals of Statistics}, \textbf{37}, 3330--3361.
\newblock Times Cited: 8.

\bibitem[{Eidsvik et~al.(2014)Eidsvik, Shaby, Reich, Wheeler and
  Niemi}]{eidsvik_estimation_2014}
Eidsvik, J., Shaby, B.~A., Reich, B.~J., Wheeler, M. and Niemi, J. (2014)
  Estimation and prediction in spatial models with block composite likelihoods.
\newblock \textit{Journal of Computational and Graphical Statistics},
  \textbf{23}, 295--315.
\newblock Publisher: Taylor \& Francis.

\bibitem[{Finley et~al.(2009)Finley, Sang, Banerjee and
  Gelfand}]{finley_improving_2009}
Finley, A.~O., Sang, H., Banerjee, S. and Gelfand, A.~E. (2009) Improving the
  performance of predictive process modeling for large datasets.
\newblock \textit{Computational Statistics \& Data Analysis}, \textbf{53},
  2873--2884.
\newblock Publisher: Elsevier.

\bibitem[{Furrer et~al.(2006)Furrer, Genton and
  Nychka}]{furrer_covariance_2006}
Furrer, R., Genton, M.~G. and Nychka, D. (2006) Covariance {Tapering} for
  {Interpolation} of {Large} {Spatial} {Datasets}.
\newblock \textit{Journal of Computational and Graphical Statistics},
  \textbf{15}, 502--523.
\newblock \urlprefix\url{https://doi.org/10.1198/106186006X132178}.
\newblock Publisher: Taylor \& Francis \_eprint:
  https://doi.org/10.1198/106186006X132178.

\bibitem[{Ghanem and Spanos(1991)}]{ghanem_stochastic_1991}
Ghanem, R.~G. and Spanos, P.~D. (1991) \textit{Stochastic {Finite} {Elements}:
  {A} {Spectral} {Approach}}.
\newblock New York: Springer New York.

\bibitem[{Golub and Underwood(1977)}]{golub_block_1977}
Golub, G. and Underwood, R. (1977) The block {Lanczos} method for computing
  eigenvalues.
\newblock \textit{Mathematical Software}, \textbf{3}, 361--377.

\bibitem[{Golub and Loan(2012)}]{golub_matrix_2012}
Golub, G.~H. and Loan, C. F.~V. (2012) \textit{Matrix {Computations}}.
\newblock Baltimore, Maryland: Johns Hopkins University Press, 4 edn.

\bibitem[{Guinness(2018)}]{guinness_permutation_2018}
Guinness, J. (2018) Permutation and {Grouping} {Methods} for {Sharpening}
  {Gaussian} {Process} {Approximations}.
\newblock \textit{Technometrics}, \textbf{60}, 415--429.
\newblock \urlprefix\url{https://doi.org/10.1080/00401706.2018.1437476}.
\newblock Publisher: Taylor \& Francis \_eprint:
  https://doi.org/10.1080/00401706.2018.1437476.

\bibitem[{Halko et~al.(2011)Halko, Martinsson and Tropp}]{halko_finding_2011}
Halko, N., Martinsson, P.~G. and Tropp, J.~A. (2011) Finding {Structure} with
  {Randomness}: {Probabilistic} {Algorithms} for {Constructing} {Approximate}
  {Matrix} {Decompositions}.
\newblock \textit{SIAM Review}, \textbf{53}, 217--288.
\newblock \urlprefix\url{https://doi.org/10.1137/090771806}.
\newblock \_eprint: https://doi.org/10.1137/090771806.

\bibitem[{Heaton et~al.(2019)Heaton, Datta, Finley, Furrer, Guinness,
  Guhaniyogi, Gerber, Gramacy, Hammerling, Katzfuss, Lindgren, Nychka, Sun and
  Zammit-Mangion}]{heaton_case_2019}
Heaton, M.~J., Datta, A., Finley, A.~O., Furrer, R., Guinness, J., Guhaniyogi,
  R., Gerber, F., Gramacy, R.~B., Hammerling, D., Katzfuss, M., Lindgren, F.,
  Nychka, D.~W., Sun, F. and Zammit-Mangion, A. (2019) A {Case} {Study}
  {Competition} {Among} {Methods} for {Analyzing} {Large} {Spatial} {Data}.
\newblock \textit{Journal of Agricultural, Biological and Environmental
  Statistics}, \textbf{24}, 398--425.
\newblock \urlprefix\url{https://doi.org/10.1007/s13253-018-00348-w}.

\bibitem[{Higdon(2002)}]{higdon_space_2002}
Higdon, D. (2002) Space and {Space}-{Time} {Modeling} using {Process}
  {Convolutions}.
\newblock In \textit{Quantitative {Methods} for {Current} {Environmental}
  {Issues}} (eds. C.~W. Anderson, V.~Barnett, P.~C. Chatwin and A.~H.
  El-Shaarawi), 37--56. London: Springer.

\bibitem[{Jacobs et~al.(2001)Jacobs, Barron and Rhodes}]{jacobs_mesoscale_2001}
Jacobs, G.~A., Barron, C.~N. and Rhodes, R.~C. (2001) Mesoscale
  characteristics.
\newblock \textit{Journal of Geophysical Research: Oceans}, \textbf{106},
  19581--19595.
\newblock \urlprefix\url{https://doi.org/10.1029/2000JC000669}.
\newblock Publisher: John Wiley \& Sons, Ltd.

\bibitem[{Jia and Liao(2009)}]{jia_accurate_2009}
Jia, L. and Liao, S. (2009) Accurate {Probabilistic} {Error} {Bound} for
  {Eigenvalues} of {Kernel} {Matrix}.
\newblock In \textit{Advances in {Machine} {Learning}} (eds. Z.-H. Zhou and
  T.~Washio), 162--175. Berlin, Heidelberg: Springer Berlin Heidelberg.

\bibitem[{Kang and Cressie(2011)}]{kang_bayesian_2011}
Kang, E.~L. and Cressie, N. (2011) Bayesian inference for the spatial random
  effects model.
\newblock \textit{Journal of the American Statistical Association},
  \textbf{106}, 972--983.
\newblock Publisher: Taylor \& Francis.

\bibitem[{Katzfuss and Cressie(2011)}]{katzfuss_spatio-temporal_2011}
Katzfuss, M. and Cressie, N. (2011) Spatio-temporal smoothing and {EM}
  estimation for massive remote-sensing data sets.
\newblock \textit{Journal of Time Series Analysis}, \textbf{32}, 430--446.
\newblock Publisher: Wiley Online Library.

\bibitem[{Kaufman et~al.(2008)Kaufman, Schervish and
  Nychka}]{kaufman_covariance_2008}
Kaufman, C.~G., Schervish, M.~J. and Nychka, D.~W. (2008) Covariance tapering
  for likelihood-based estimation in large spatial data sets.
\newblock \textit{Journal of the American Statistical Association},
  \textbf{103}, 1545--1555.
\newblock Publisher: Taylor \& Francis.

\bibitem[{Kennedy and O'Hagan(2001)}]{kennedy_bayesian_2001}
Kennedy, M.~C. and O'Hagan, A. (2001) Bayesian {Calibration} of {Computer}
  {Models}.
\newblock \textit{Journal of the Royal Statistical Society Series B:
  Statistical Methodology}, \textbf{63}, 425--464.
\newblock \urlprefix\url{https://doi.org/10.1111/1467-9868.00294}.

\bibitem[{Koltchinskii and Giné(2000)}]{koltchinskii_random_2000}
Koltchinskii, V. and Giné, E. (2000) Random matrix approximation of spectra of
  integral operators.
\newblock \textit{Bernoulli}, \textbf{6}, 113--167.
\newblock
  \urlprefix\url{https://projecteuclid.org/journals/bernoulli/volume-6/issue-1/Random-matrix-approximation-of-spectra-of-integral-operators/bj/1082665383.full}.
\newblock Publisher: Bernoulli Society for Mathematical Statistics and
  Probability.

\bibitem[{König(1986)}]{konig_eigenvalue_1986}
König, H. (1986) \textit{Eigenvalue {Distribution} of {Compact} {Operators}},
  vol.~16 of \textit{Operator {Theory}: {Advances} and {Applications}}.
\newblock Basel: Birkhäuser Verlag.

\bibitem[{Lemos and Sansó(2009)}]{lemos_spatio-temporal_2009}
Lemos, R.~T. and Sansó, B. (2009) A spatio-temporal model for mean, anomaly,
  and trend fields of {North} {Atlantic} sea surface temperature.
\newblock \textit{Journal of the American Statistical Association},
  \textbf{104}, 5--18.
\newblock Publisher: Taylor \& Francis.

\bibitem[{Lindgren et~al.(2011)Lindgren, Rue and
  Lindström}]{lindgren_explicit_2011}
Lindgren, F., Rue, H. and Lindström, J. (2011) An explicit link between
  {Gaussian} fields and {Gaussian} {Markov} random fields: the stochastic
  partial differential equation approach.
\newblock \textit{Journal of the Royal Statistical Society: Series B
  (Statistical Methodology)}, \textbf{73}, 423--498.
\newblock
  \urlprefix\url{https://onlinelibrary.wiley.com/doi/abs/10.1111/j.1467-9868.2011.00777.x}.

\bibitem[{Long(2021)}]{long_aquatic_2021}
Long, M.~H. (2021) Aquatic {Biogeochemical} {Eddy} {Covariance} {Fluxes} in the
  {Presence} of {Waves}.
\newblock \textit{Journal of Geophysical Research: Oceans}, \textbf{126},
  e2020JC016637.
\newblock \urlprefix\url{https://doi.org/10.1029/2020JC016637}.
\newblock Publisher: John Wiley \& Sons, Ltd.

\bibitem[{Musco and Musco(2015)}]{musco_randomized_2015}
Musco, C. and Musco, C. (2015) Randomized {Block} {Krylov} {Methods} for
  {Stronger} and {Faster} {Approximate} {Singular} {Value} {Decomposition}.
\newblock In \textit{Advances in {Neural} {Information} {Processing} {Systems}}
  (eds. C.~Cortes, N.~Lawrence, D.~Lee, M.~Sugiyama and R.~Garnett), vol.~28.
  Curran Associates, Inc.
\newblock
  \urlprefix\url{https://proceedings.neurips.cc/paper/2015/file/1efa39bcaec6f3900149160693694536-Paper.pdf}.

\bibitem[{Nychka et~al.(2015)Nychka, Bandyopadhyay, Hammerling, Lindgren and
  Sain}]{nychka_multiresolution_2015}
Nychka, D., Bandyopadhyay, S., Hammerling, D., Lindgren, F. and Sain, S. (2015)
  A {Multiresolution} {Gaussian} {Process} {Model} for the {Analysis} of
  {Large} {Spatial} {Datasets}.
\newblock \textit{Journal of Computational and Graphical Statistics},
  \textbf{24}, 579--599.
\newblock \urlprefix\url{http://www.jstor.org/stable/24737282}.
\newblock Publisher: [American Statistical Association, Taylor \& Francis,
  Ltd., Institute of Mathematical Statistics, Interface Foundation of America].

\bibitem[{O'Hagan(1978)}]{ohagan_curve_1978}
O'Hagan, A. (1978) Curve fitting and optimal design for predictions.
\newblock \textit{Journal of the Royal Statistical Society B}, \textbf{40},
  1--42.

\bibitem[{Pietsch(1987)}]{pietsch_eigenvalues_1987}
Pietsch, A. (1987) \textit{Eigenvalues and s-{Numbers}}, vol.~13 of
  \textit{Cambridge {Studies} in {Advanced} {Mathematics}}.
\newblock Cambridge: Cambridge University Press.

\bibitem[{Rasmussen and Williams(2006)}]{rasmussen_gaussian_2006}
Rasmussen, C.~E. and Williams, C. K.~I. (2006) \textit{Gaussian {Processes} for
  {Machine} {Learning}}.
\newblock MIT Press.
\newblock \urlprefix\url{http://www.gaussianprocess.org/gpml/}.

\bibitem[{Raykar and Duraiswami(2007)}]{raykar_kernel_2007}
Raykar, V.~C. and Duraiswami, R. (2007) Kernel methods in signal processing.
\newblock \textit{Signal Processing Magazine, IEEE}, \textbf{24}, 118--121.
\newblock Publisher: IEEE.

\bibitem[{Rue et~al.(2009)Rue, Martino and Chopin}]{rue_approximate_2009}
Rue, H., Martino, S. and Chopin, N. (2009) Approximate {Bayesian} inference for
  latent {Gaussian} models by using integrated nested {Laplace} approximations.
\newblock \textit{Journal of the Royal Statistical Society: Series B
  (Statistical Methodology)}, \textbf{71}, 319--392.
\newblock Publisher: Wiley Online Library.

\bibitem[{Santin and Schaback(2016)}]{santin_approximation_2016}
Santin, G. and Schaback, R. (2016) Approximation of eigenfunctions in
  kernel-based spaces.
\newblock \textit{Advances in Computational Mathematics}, \textbf{42},
  973--993.
\newblock \urlprefix\url{https://doi.org/10.1007/s10444-015-9449-5}.

\bibitem[{Sarlos(2006)}]{sarlos_improved_2006}
Sarlos, T. (2006) Improved {Approximation} {Algorithms} for {Large} {Matrices}
  via {Random} {Projections}.
\newblock In \textit{2006 47th {Annual} {IEEE} {Symposium} on {Foundations} of
  {Computer} {Science} ({FOCS}'06)}, 143--152.
\newblock ISSN: 0272-5428.

\bibitem[{Schaback(1995)}]{schaback_error_1995}
Schaback, R. (1995) Error estimates and condition numbers for radial basis
  function interpolation.
\newblock \textit{Advances in Computational Mathematics}, \textbf{3}, 251--264.
\newblock \urlprefix\url{https://doi.org/10.1007/BF02432002}.

\bibitem[{Schaback and Wendland(2002)}]{buhmann_approximation_2002}
Schaback, R. and Wendland, H. (2002) Approximation by positive definite
  kernels.
\newblock In \textit{Advanced {Problems} in {Constructive} {Approximation}}
  (eds. M.~D. Buhmann and D.~H. Mache), 203--222. Basel: Birkhäuser Basel.

\bibitem[{Schölkopf and Smola(2002)}]{scholkopf_learning_2002}
Schölkopf, B. and Smola, A.~J. (2002) \textit{Learning with {Kernels}}.
\newblock Cambridge, MA: MIT Press.

\bibitem[{Schölkopf et~al.(2004)Schölkopf, Tsuda and
  Vert}]{scholkopf_kernel_2004}
Schölkopf, B., Tsuda, K. and Vert, J.-P. (eds.) (2004) \textit{Kernel methods
  in computational biology}.
\newblock MIT Press.

\bibitem[{Shawe-Taylor and Cristianini(2004)}]{shawe-taylor_kernel_2004}
Shawe-Taylor, J. and Cristianini, N. (2004) \textit{Kernel methods for pattern
  analysis}.
\newblock Cambridge University Press.

\bibitem[{Stein(2014)}]{stein_limitations_2014}
Stein, M.~L. (2014) Limitations on low rank approximations for covariance
  matrices of spatial data.
\newblock \textit{Spatial Statistics}, \textbf{8}, 1--19.
\newblock
  \urlprefix\url{https://www.sciencedirect.com/science/article/pii/S2211675313000390}.

\bibitem[{Stein et~al.(2013)Stein, Chen, Anitescu and
  al}]{stein_stochastic_2013}
Stein, M.~L., Chen, J., Anitescu, M. and al, e. (2013) Stochastic approximation
  of score functions for {Gaussian} processes.
\newblock \textit{The Annals of Applied Statistics}, \textbf{7}, 1162--1191.
\newblock Publisher: Institute of Mathematical Statistics.

\bibitem[{Stein et~al.(2004)Stein, Chi and Welty}]{stein_approximating_2004}
Stein, M.~L., Chi, Z. and Welty, L.~J. (2004) Approximating likelihoods for
  large spatial data sets.
\newblock \textit{Journal of the Royal Statistical Society: Series B
  (Statistical Methodology)}, \textbf{66}, 275--296.
\newblock Publisher: Wiley.

\bibitem[{Stroud et~al.(2017)Stroud, Stein and Lysen}]{stroud_bayesian_2017}
Stroud, J.~R., Stein, M.~L. and Lysen, S. (2017) Bayesian and {Maximum}
  {Likelihood} {Estimation} for {Gaussian} {Processes} on an {Incomplete}
  {Lattice}.
\newblock \textit{Journal of Computational and Graphical Statistics},
  \textbf{26}, 108--120.
\newblock \urlprefix\url{https://doi.org/10.1080/10618600.2016.1152970}.
\newblock Publisher: Taylor \& Francis \_eprint:
  https://doi.org/10.1080/10618600.2016.1152970.

\bibitem[{Sun et~al.(2012)Sun, Li and Genton}]{sun_geostatistics_2012}
Sun, Y., Li, B. and Genton, M.~G. (2012) Geostatistics for {Large} {Datasets}.
\newblock In \textit{Advances and {Challenges} in {Space}-time {Modelling} of
  {Natural} {Events}} (eds. E.~Porcu, J.~Montero and M.~Schlather), Lecture
  {Notes} in {Statistics}, 55--77. Berlin, Heidelberg: Springer.

\bibitem[{Tang et~al.(2021)Tang, Zhang and
  Banerjee}]{tang_identifiability_2021}
Tang, W., Zhang, L. and Banerjee, S. (2021) On identifiability and consistency
  of the nugget in {Gaussian} spatial process models.
\newblock \textit{Journal of the Royal Statistical Society: Series B
  (Statistical Methodology)}, \textbf{83}, 1044--1070.
\newblock
  \urlprefix\url{https://onlinelibrary.wiley.com/doi/abs/10.1111/rssb.12472}.
\newblock \_eprint: https://onlinelibrary.wiley.com/doi/pdf/10.1111/rssb.12472.

\bibitem[{Wahba(2002)}]{wahba_soft_2002}
Wahba, G. (2002) Soft and hard classification by reproducing kernel {Hilbert}
  space methods.
\newblock \textit{Proceedings of the National Academy of Sciences},
  \textbf{99}, 16524--16530.
\newblock \urlprefix\url{https://www.pnas.org/doi/10.1073/pnas.242574899}.
\newblock Publisher: Proceedings of the National Academy of Sciences.

\bibitem[{Wang et~al.(2023)Wang, Dahl, Swersky, Lee, Nado, Gilmer, Snoek and
  Ghahramani}]{wang_pre-trained_2023}
Wang, Z., Dahl, G.~E., Swersky, K., Lee, C., Nado, Z., Gilmer, J., Snoek, J.
  and Ghahramani, Z. (2023) Pre-trained {Gaussian} processes for {Bayesian}
  optimization.
\newblock \urlprefix\url{http://arxiv.org/abs/2109.08215}.
\newblock ArXiv:2109.08215 [cs, stat].

\bibitem[{Williams and Seeger(2000)}]{williams_using_2000}
Williams, C. and Seeger, M. (2000) Using the {Nyström} {Method} to {Speed}
  {Up} {Kernel} {Machines}.
\newblock In \textit{Advances in {Neural} {Information} {Processing} {Systems}}
  (eds. T.~Leen, T.~Dietterich and V.~Tresp), vol.~13. MIT Press.

\bibitem[{Xu and Gong(2003)}]{xu_background_2003}
Xu, Q. and Gong, J. (2003) Background error covariance functions for {Doppler}
  radial-wind analysis.
\newblock \textit{Quarterly Journal of the Royal Meteorological Society},
  \textbf{129}, 1703--1720.
\newblock \urlprefix\url{https://doi.org/10.1256/qj.02.129}.
\newblock Publisher: John Wiley \& Sons, Ltd.

\end{thebibliography}
